\documentclass{article} 
\usepackage{iclr2026_conference,times}

\usepackage{amsmath,amsfonts,bm}

\def\eqref#1{equation~\ref{#1}}

\def\1{\bm{1}}

\DeclareMathAlphabet{\mathsfit}{\encodingdefault}{\sfdefault}{m}{sl}
\SetMathAlphabet{\mathsfit}{bold}{\encodingdefault}{\sfdefault}{bx}{n}

\usepackage{hyperref}
\usepackage{url}

\title{Hidden Meanings in Plain Sight: RebusBench for Evaluating Cognitive Visual Reasoning}

\author{Seyed Amir Kasaei, Arash Marioriyad, Mahbod Khaleti, \\
\textbf{MohammadAmin Fazli, Mahdieh Soleymani Baghshah \& Mohammad Hossein Rohban} \\
Department of Computer Engineering\\
Sharif University of Technology\\
\texttt{a.kasaei@me.com}, \texttt{\{arashmarioriyad, mahbod.kh2005\}@gmail.com} \\
\texttt{\{fazli, soleymani, rohban\}@sharif.edu}}

\usepackage{pifont}
\usepackage{adjustbox}  
\usepackage{booktabs}   
\usepackage{multirow}
\usepackage[table]{xcolor}
\definecolor{tablegray}{gray}{0.95}
\definecolor{gptcol}{RGB}{240,248,255} 

\usepackage[most]{tcolorbox} 
\usepackage{xcolor}
\usepackage{graphicx}

\definecolor{lvlmframe}{RGB}{50, 80, 140}   
\definecolor{lvlmback}{RGB}{240, 245, 250}  

\definecolor{evalframe}{RGB}{40, 120, 80}   
\definecolor{evalback}{RGB}{240, 250, 240}  

\newtcolorbox{lvlmprompt}[2][]{
    enhanced,
    title={#2}, 
    fonttitle=\bfseries\sffamily,
    coltitle=white,
    colframe=lvlmframe,
    colback=lvlmback,
    attach boxed title to top left={xshift=4mm, yshift=-3mm},
    boxed title style={
        colback=lvlmframe,
        arc=2mm,
        outer arc=2mm,
        boxrule=0pt,
        drop fuzzy shadow
    },
    arc=3mm,
    boxrule=0.5pt,
    leftrule=4mm, 
    drop fuzzy shadow, 
    fontupper=\small\ttfamily\color{black!85}, 
    top=12pt, 
    #1
}

\newtcolorbox{llmevalbox}[2][]{
    enhanced,
    title={#2}, 
    fonttitle=\bfseries\sffamily,
    coltitle=white,
    colframe=evalframe,
    colback=evalback,
    attach boxed title to top left={xshift=4mm, yshift=-3mm},
    boxed title style={
        colback=evalframe,
        arc=2mm,
        outer arc=2mm,
        boxrule=0pt,
        drop fuzzy shadow
    },
    arc=3mm,
    boxrule=0.5pt,
    leftrule=4mm, 
    drop fuzzy shadow,
    fontupper=\small\ttfamily\color{black!85},
    top=12pt,
    #1
}

\iclrfinalcopy 
\begin{document}

\maketitle

\begin{abstract}
Large Vision--Language Models (LVLMs) have achieved remarkable proficiency in explicit visual recognition, effectively describing what is directly visible in an image. However, a critical cognitive gap emerges when the visual input serves only as a clue rather than the answer. We identify that current models struggle with the complex, multi-step reasoning required to solve problems where information is not explicitly depicted. Successfully solving a rebus puzzle requires a distinct cognitive workflow: the model must extract visual and textual attributes, retrieve linguistic prior knowledge (such as idioms), and perform abstract mapping to synthesize these elements into a meaning that exists outside the pixel space. To evaluate this neurosymbolic capability, we introduce \textbf{RebusBench}, a benchmark of 1,164 puzzles designed to test this specific integration of perception and knowledge. Our evaluation of state-of-the-art models (including Qwen, InternVL, and LLaVA) shows a severe deficiency: performance saturates below 10\% Exact Match and 20\% semantic accuracy, with no significant improvement observed from model scaling or In-Context Learning (ICL). These findings suggest that while models possess the necessary visual and linguistic components, they lack the cognitive reasoning ``glue'' to connect them. Project page available at \href{https://amirkasaei.com/rebusbench/}{this URL}
\end{abstract}
\section{Introduction}

Visual reasoning extends beyond perception, requiring the synthesis of visual inputs with abstract knowledge. Large Vision--Language Models (LVLMs) have recently demonstrated remarkable progress in this domain. Systems ranging from foundational architectures like Flamingo and BLIP \cite{flamingo, blip, blip2} to recent instruction-tuned and scaled models such as LLaVA, InternVL, and Qwen-VL \cite{llava, instructblip, chen2024internvl, qwen2-5, qwen3} have set new standards in visual question answering and spatial grounding. Together with proprietary models like GPT-5.2~\cite{gpt52} and Gemini 3~\cite{gemini3}, these advancements suggest that modern LVLMs are increasingly capable of bridging the gap between pixel-level processing and semantic understanding.

The advancement of the field relies heavily on the quality of its benchmarks, yet existing evaluation suites often fail to probe the cognitive depth of these models. Standard datasets such as VQA v2, GQA, and CLEVR \cite{antol2015vqa, hudson2019gqa, johnson2017clevr} predominantly assess referential grounding—the capacity to map linguistic tokens to explicit visual features. From a cognitive perspective, this mirrors rapid, ``System 1'' perceptual processing rather than deep, deliberate reasoning. These tasks rarely require the model to engage in modal entanglement, where visual and textual elements must be creatively recombined to form new concepts. Consequently, current benchmarks struggle to differentiate between shallow pattern matching and the multi-step, open-ended symbolic manipulation required for true human-level intelligence.

\begin{figure*}[t]
    \centering
    \includegraphics[width=\textwidth]{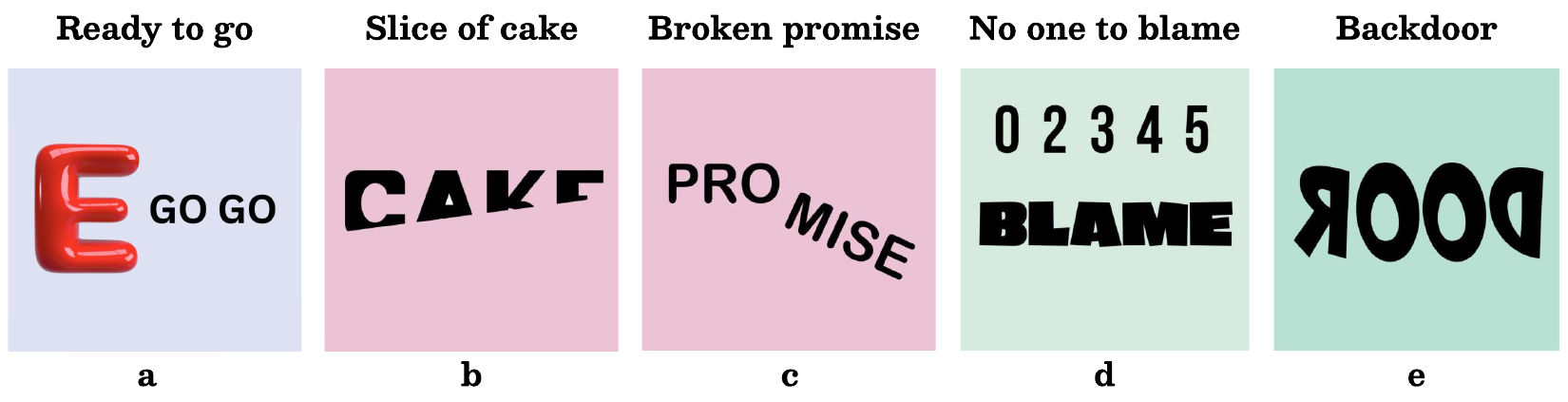}
    \caption{\textbf{Sample puzzles from RebusBench.} Solving these requires transcending literal perception ("System 1") to perform abstract ``System 2'' reasoning, where models must entangle visual cues (position, color, style) with linguistic knowledge to reconstruct the hidden idiom.}
    \label{fig:teaser}
\end{figure*}

To this end, we introduce \textbf{RebusBench}, a novel benchmark based on rebus puzzles—visual riddles that serve as a rigorous testbed for deep, cognitively-inspired reasoning. Unlike standard tasks where the answer is explicitly present in the image, a rebus puzzle demands a constructive inference process. As illustrated in Figure~\ref{fig:teaser}, consider a puzzle displaying a red letter ``E'' alongside two instances of the word ``GO''. To arrive at the correct solution, ``\textit{Ready to go},'' a model cannot merely describe the visible elements. It must execute a complex cognitive chain: (i) extract the visual attributes (color red, repetition of ``GO''); (ii) retrieve external phonological and semantic knowledge (mapping ``Red E'' $\rightarrow$ ``Ready'' and ``Two Gos'' $\rightarrow$ ``to go''); and (iii) synthesize these components into a coherent idiomatic phrase. This setting forces the model to move beyond simple perception and engage in visual--textual entanglement, distinguishing generic recognition from the ability to perform rigorous, multi-step symbolic reasoning.

Our empirical evaluation reveals a stark reality: state-of-the-art LVLMs struggle profoundly with this benchmark. Even the largest open-weight models, such as Qwen 2.5-32B and InternVL-30B, fail to surpass 10\% in Exact Match (EM) accuracy. Crucially, standard mitigation strategies offer little relief; neither scaling model parameters nor increasing context through few-shot prompting yields significant improvements. Furthermore, even when evaluated by a lenient semantic judge (GPT-4o) to account for near-synonyms, performance plateaus below 20\% across all settings. This resistance to scaling and In-Context Learning (ICL) underscores a fundamental limitation in current architectures: they lack the robust vision--text entanglement necessary for abstract reasoning, suggesting that raw compute and data are insufficient to bridge this specific cognitive gap.

\section{Related Work}

\paragraph{Scene Understanding and Image Captioning.} 
Foundational datasets such as MS COCO \cite{lin2014microsoft}, Flickr30k \cite{young2014from}, Visual Genome \cite{krishna2017visual}, and LVIS \cite{gupta2019lvis} emphasize literal mappings between visual regions and linguistic labels. While knowledge-augmented variants like OK-VQA \cite{marino2019ok} incorporate external facts, they treat vision and text as parallel streams for retrieval rather than entangled components. These frameworks largely overlook scenarios where the visual rendering of text or the symbolic fusion of modalities necessitates non-literal interpretation and decoding of hidden semantic identities.

\paragraph{Spatial and Compositional Reasoning.} Benchmarks including CLEVR \cite{johnson2017clevr}, GQA \cite{hudson2019gqa}, VSR \cite{liu2023visual}, SpatialVLM \cite{chen2024spatialvlm}, Spatial4D-Bench \cite{wang2026spatial4d}, iVISPAR \cite{mayer2025ivispar}, and SpatialBench \cite{xu2025spatialbench} evaluate a model's grasp of geometric coordinates and relative 3D positioning. However, these tasks typically view spatiality through a physical lens rather than a symbolic one. They fail to assess structural arrangement as a syntactic operator—where the placement of elements (e.g., vertical stacking or specific alignments) transforms individual components into a unified abstract concept rather than a mere physical description.


\paragraph{Visual Counting and Quantity Estimation.} 
Tasks focused on enumeration, such as Count7W \cite{chattopadhyay2017counting}, TallyQA \cite{acharya2019tallyqa}, HowManyQA \cite{trott2020howmany}, and CAPTURe \cite{pothiraj2025capture}, measure precise object individuation. In these settings, cardinality is typically the terminal objective. Existing benchmarks lack requirements for multi-step synthesis, where numerical results must serve as intermediate symbolic clues. Consequently, they do not test a model’s ability to creatively integrate counts with other visual-textual cues to resolve open-ended or latent riddles.

\paragraph{Advanced Reasoning and Multi-discipline Benchmarks.} 
Comprehensive evaluative sets like ScienceQA \cite{lu2022learn}, MMMU \cite{yue2023mmmu}, MathVista \cite{lu2023mathvista}, and SEED-Bench \cite{li2023seed} stress expert-level knowledge and formulaic deduction. While benchmarks in this category test general intelligence, they remain grounded in linear logic and factual recall. There remains a significant gap in evaluating lateral thinking and "out-of-the-box" creativity, particularly in resolving ill-posed visual problems that lack a direct, deductive path to a solution.

\section{The RebusBench Dataset}

To rigorously evaluate abstract visual reasoning, we introduce \textbf{RebusBench}, a dataset of \textbf{1,164 rebus puzzles} aggregated from diverse sources specializing in lateral thinking \cite{eslvault_rebus_puzzles, justfamilyfun_rebus_puzzles_2025}. In contrast to synthetic VQA datasets that often rely on rigid, procedurally generated patterns, RebusBench features human-authored puzzles grounded in authentic idiomatic usage. We collected both the puzzle images and their intended solutions directly from mentioned sources. Subsequently, we applied strict manual verification to filter out ambiguities, ensuring that every retained instance possesses a unique, clear solution free from obscure cultural trivia or subjective interpretation.

A defining feature of RebusBench is that it eschews a fixed ratio of visual-to-textual reasoning in favor of a continuous cognitive spectrum. On one end, the dataset includes \textbf{visually dominant} instances driven by geometric deformation, where the physical manipulation of text constitutes the primary clue. For example, in the \textit{``Slice of cake''} puzzle (Figure~\ref{fig:teaser}b), the word ``CAKE'' is visually rendered with a slice cut out of it. Solving this requires the model to treat the text as a malleable object rather than a fixed string, mapping the missing slice to the concept of ``slicing.'' Conversely, the benchmark covers \textbf{symbolically dominant} puzzles that rely on abstract logical patterns and absence. In the \textit{``No one to blame''} puzzle (Figure~\ref{fig:teaser}d), the image displays a numerical sequence ``0, 2, 3...'' alongside the word ``BLAME.'' Here, the reasoning is driven by the specific absence of the number ``1''; the model must detect this void, map it to the linguistic concept ``No one,'' and concatenate it with the visible text. This diversity ensures that models are tested on their ability to dynamically weight visual attributes and symbolic sequences rather than relying on a single modality.

We formalize this task as open-ended generative reasoning. Given an image $I$, the model must generate the target idiom $T$. Theoretically, this requires the model to approximate a function $f(I) \rightarrow T$ that implicitly models a reasoning chain $R$, entangling visual attributes $V$ (position, color, size) with linguistic priors $L$ (phonology, idioms). Crucially, this is a \textit{suppressive} task: to generate the correct target $T$ (e.g., ``Ready to go''), the model must suppress the high-probability literal caption (e.g., ``A red letter E next to two Gos'')—a hallmark of System 2 cognitive control.

\begin{table}[hb]
\centering
\caption{\textbf{Quantitative Performance on RebusBench.} We report Exact Match (EM) and GPT-4o Semantic Judge scores across One-shot and Three-shot settings. The results highlight a universal struggle across all architectures, where neither model scaling nor few-shot prompting yields significant improvements, confirming a fundamental gap in abstract visual reasoning.}
\vspace{0.3cm}
\label{tab:main_results}
\resizebox{0.65\textwidth}{!}{%
\begin{tabular}{l|cccc}
\toprule
\textbf{MODEL} &
\multicolumn{2}{c}{\textbf{1-SHOT}} &
\multicolumn{2}{c}{\textbf{3-SHOT}} \\
\cmidrule(lr){2-3}
\cmidrule(lr){4-5}
 & \textbf{EM (\% $\uparrow$)} & \textbf{GPT-4O ($\uparrow$)}
 & \textbf{EM (\% $\uparrow$)} & \textbf{GPT-4O ($\uparrow$)} \\
\midrule
LLaVA-1.5 7B      & 1.20   & 0.1491      & 1.03 & 0.1381 \\
\midrule
InternVL 3.5 4B   & 4.73 & 0.1383 & 4.04 & 0.1272 \\
InternVL 3.5 8B   & 4.81 & 0.1397 & 4.64 & 0.1295 \\
InternVL 3.5 30B  & 5.15 & 0.1430 & 6.36 & 0.1462 \\
\midrule
Qwen 2.5 3B       & 3.52 & 0.1351 & 4.47 & 0.1340 \\
Qwen 2.5 7B       & 5.15 & 0.1413 & 6.10 & 0.1496 \\
Qwen 2.5 32B      & 7.39 & 0.1478 & 8.08 & 0.1508 \\
\midrule
Qwen 3 4B         & 4.12 & 0.1650 & 5.84 & 0.1638 \\
Qwen 3 8B         & 6.36 & 0.1661 & 6.36 & 0.1704 \\
\bottomrule
\end{tabular}%
}
\end{table}

\section{Experiments}

\subsection{Experimental Setup}

\paragraph{Models.} We evaluate a representative suite of open-weight LVLMs, varying in model families and parameter scales to benchmark current capabilities. Our evaluation includes LLaVA-1.5 7B \cite{llava}, alongside the more recent InternVL 3.5 series (4B, 8B, 30B) \cite{chen2024internvl}, Qwen 2.5 series (3B, 7B, 32B) \cite{qwen2-5}, and the Qwen 3 series (4B, 8B) \cite{qwen3}.

\paragraph{Prompting Strategies.} We assess performance across One-shot and Three-shot In-Context Learning (ICL) settings. In these few-shot configurations, we provide solved examples explained in the textual input prompt to guide the reasoning process. The full prompts for each configuration are provided in Appendix~\ref{sec:lvlm}.

\paragraph{Evaluation Metrics.} We employ two complementary metrics. We report \textbf{Exact Match (EM)}, a strict metric where predictions are normalized (lowercased, stripped of whitespace/special characters) before comparison. To account for linguistic variations, we also use a \textbf{GPT-4o Judge (Semantic Score)}~\cite{gpt4o}, which rates semantic similarity on a continuous scale from 0.0 to 1.0. The full evaluation prompt is detailed in Appendix~\ref{sec:llm}.

\subsection{Results}

\paragraph{Overall Performance and Model Scaling.} As summarized in Table~\ref{tab:main_results}, the evaluation reveals a distinct performance ceiling across all model families. We observe a universal failure to reliably solve rebus puzzles, with Exact Match (EM) scores saturating below $10\%$ regardless of the architecture. Notably, increasing model parameters offers negligible returns. Transitions from smaller to significantly larger variants within the same family yield only marginal gains. For instance, scaling the InternVL 3.5 architecture from 4B to 30B only improves 1-shot performance from $4.73\%$ to $5.15\%$. Similarly, the massive Qwen 2.5 32B model ($7.39\%$) shows limited improvement over its 7B counterpart ($5.15\%$) in the 1-shot setting. Even with 3-shot prompting, performance remains stagnant, with the best model achieving only $8.08\%$. This plateau suggests that simply adding capacity does not enable the models to spontaneously emerge the ability to bridge visual perception and idiomatic abstraction.

\paragraph{Impact of In-Context Learning.} Furthermore, we find that In-Context Learning (ICL) remains insufficient for bridging this reasoning gap. Surprisingly, increasing the number of solved examples often yields negligible gains or even degrades performance. For instance, the InternVL 3.5 4B model drops from $4.73\%$ (1-shot) to $4.04\%$ (3-shot), while the Qwen 3 8B model remains exactly stagnant at $6.36\%$ across both settings. Even the largest model, Qwen 2.5 32B, sees only a marginal increase from $7.39\%$ to $8.08\%$. When evaluated by the more lenient GPT-4o semantic judge, performance remains universally low, peaking at only $0.1704$ (Qwen 3 8B, 3-shot). This resistance to few-shot prompting demonstrates that current models cannot easily induce the necessary ``System 2'' logic from demonstrations alone.

\section{Conclusion and Future Work}

\paragraph{Conclusion.} We introduced \textbf{RebusBench} to evaluate the neurosymbolic reasoning of LVLMs. Our experiments reveal a critical performance ceiling: regardless of model scale or few-shot prompting, architectures consistently fail to surpass 10\% Exact Match and 20\% semantic accuracy. These results highlight a fundamental deficiency in current models, which struggle to entangle visual perception with linguistic abstraction—a necessary step for ``System 2'' reasoning.

\paragraph{Future Work.} Future efforts will focus on expanding the dataset's diversity and adding fine-grained metadata to classify puzzles along the visual--textual spectrum. This will enable precise diagnosis of modality biases. Additionally, we plan to release a standardized evaluation pipeline and extend our benchmarking to include proprietary closed-source models, establishing a comprehensive baseline for the field.

\clearpage
\newpage

\bibliography{ref}
\bibliographystyle{iclr2026_conference}

\clearpage
\newpage
\appendix
\textbf{\Large Appendix}

\section{Prompting Strategy}\label{sec:prompt}

\subsection{Assessing the Model (LVLM Inference)}\label{sec:lvlm}
To benchmark the visual reasoning capabilities of the LVLM, we employ prompt variations aimed at solving the rebus puzzles. Specifically, we utilize \textbf{One-Shot \& Three-Shot} settings where we provide one or three visual-text pairs as context. This approach evaluates the model's capacity for in-context learning, testing whether it can generalize the pattern of extracting semantic meaning from visual layout based on the provided examples.

\begin{lvlmprompt}{LVLM Input: One-Shot Rebus Solver}
"You are given an image that represents a rebus puzzle (a visual word riddle).\\
A rebus puzzle encodes a common English word or phrase using visual layout, repetition, color, position, or size of text and symbols.\\
Do NOT read the image literally.\\
Instead, infer the hidden word or idiomatic expression suggested by the visual arrangement.\\\\
\textbf{Example:}\\
- A red letter 'E' followed by 'GO GO' means 'ready to go'.\\\\
\textbf{Question:} What English word or phrase is represented?\\
Return ONLY the final answer in 1-5 words.\\
Do not explain."
\end{lvlmprompt}

\begin{lvlmprompt}{LVLM Input: Three-Shot Rebus Solver}
"You are given an image that represents a rebus puzzle (a visual word riddle).\\
A rebus puzzle encodes a common English word or phrase using visual layout, repetition, color, position, or size of text and symbols.\\
Do NOT read the image literally.\\
Instead, infer the hidden word or idiomatic expression suggested by the visual arrangement.\\\\
\textbf{Examples:}\\
- The word 'MAN' written three times means 'three men'.\\
- The word 'READ' placed inside a box means 'read between the lines'.\\
- A red letter 'E' followed by 'GO GO' means 'ready to go'.\\\\
\textbf{Question:} What English word or phrase is represented?\\
Return ONLY the final answer in 1-5 words.\\
Do not explain."
\end{lvlmprompt}

\subsection{Evaluating Model Output (LLM Judge)}
\label{sec:llm}
Since rebus puzzles often have synonymous answers, strict string matching is insufficient. We utilize a text-only LLM as a semantic judge. The evaluation prompt inputs the \texttt{Ground Truth} and the LVLM's \texttt{Predicted} answer. The LLM is instructed to output a scalar score ranging from 0.0 to 1.0, penalizing unrelated answers while rewarding semantically correct interpretations.

\begin{llmevalbox}{LLM Evaluation: Rebus Judge}
You are an expert evaluator for Rebus puzzles. \\
Your task is to compare a 'Ground Truth' answer with a 'Predicted' answer.\\\\
\textbf{Ground Truth:} "\{ground\_truth\}"\\
\textbf{Predicted:} "\{prediction\}"\\\\
\textbf{Scoring Criteria:}\\
- Score 1.0: Perfect match or semantically identical (e.g., "Middle-aged" vs "middle aged", "Apple" vs "Apples").\\
- Score 0.0: Completely unrelated.\\
- Otherwise: Based on the level of capturing the core concept in ground truth and partially correctness.\\\\
Return ONLY a single numerical float between 0.0 and 1.0. \\
No explanations, no text.
\end{llmevalbox}

\end{document}